\documentclass{article}

\usepackage{arxiv}
\usepackage{graphicx}
\usepackage{xcolor}
\usepackage{amsmath}
\usepackage{caption}
\usepackage{subcaption}
\usepackage[utf8]{inputenc} 
\usepackage[T1]{fontenc}    
\usepackage{hyperref}       
\usepackage{url}            
\usepackage{booktabs}       
\usepackage{amsfonts}       
\usepackage{nicefrac}       
\usepackage{microtype}      
\usepackage{lipsum}
\usepackage{longtable}
\usepackage{multirow}
\usepackage{rotating}
\usepackage{geometry}
\usepackage{pdflscape}
\usepackage{afterpage}
\usepackage[paper=portrait,pagesize]{typearea}
\usepackage{amssymb}

\title{Plant Diseases recognition on images using Convolutional Neural Networks: A Systematic Review}

\author{André S. Abade, Paulo Afonso Ferreira and Fl{\'a}vio de Barros Vidal}

\begin{document}

\maketitle

\abstract{Plant diseases are considered one of the main factors influencing food production and minimize losses in production, and it is essential that crop diseases have fast detection and recognition.  The recent expansion of deep learning methods has found its application in plant disease detection, offering a robust tool with highly accurate results. In this context, this work presents a systematic review of the literature that aims to identify the state of the art of the use of convolutional neural networks(CNN) in the process of identification and classification of plant diseases, delimiting trends, and indicating gaps.  In this sense, we present  121 papers selected in the last ten years with different approaches to treat aspects related to disease detection, characteristics of the data set, the crops and pathogens investigated. From the results of the systematic review, it is possible to understand the innovative trends regarding the use of CNNs in the identification of plant diseases and to identify the gaps that need the attention of the research community.}

\section{Introduction}
\label{S:1}
Plant diseases are considered one of the main factors influencing food production, being responsible for the significant reduction of the physical or economic productivity of the crops and, in some instances, maybe an impediment to this activity. According to~\cite{ALTIERI:2018}, to minimize production losses and maintain crop sustainability, it is essential that disease management and control measures be carried out appropriately, highlighting the constant monitoring of the crop, combined with the rapid and accurate diagnosis of the diseases. These practices are the most recommended by phytopathologists~\cite{ALTIERI:2018}.

The major challenge for agriculture is the correct identification of the symptoms of major diseases that affect crops~\cite{Mohanty:2016}. Manual and mechanized practices in traditional planting processes are not able to cover large areas of plantation and provide essential early information to decision-making processes, according to~\cite{MILLER:2009}. Thus, it is necessary to develop {\it auTomatod} solutions, practical, reliable, and economically able to monitor the health of plants providing meaningful information to the decision-making process, for example, the application and correct dosage of pesticides in specific treatment certain diseases~\cite{MAHLEIN:2016}.

Computer Vision, along with Artificial Intelligence (AI), has been developing techniques and methods for recognizing and classifying objects with significant advances~\cite{ARNAL:2013}. These systems use Convolutional Neural Networks (CNNs)~\cite{LECUN:1998}, and their results in some experiments are already superior to humans in large-scale reconnaissance tasks. The works of~\cite{SANKARAN:2010} and~\cite{Martinelli:2015} highlight the use of images as one of the most advanced methods of detection and recognition of plant diseases.

Many approaches already make use of popular architectures such as LeNet\cite{LeNet:1998}, AlexNet\cite{AlexNet:2012}, VGGNet\cite{VGGNet:2014}, GoogLeNet\cite{GoogLeNet:2015}, InceptionV3\cite{InceptionV3:2016}, ResNet\cite{ResNet:2016} and DenseNet\cite{DenseNet:2017}, considerably increasing the accuracy in the identification of plant diseases. However, numerous challenges still hinder the correct classification of phytopathology, such as the genetic and phenotypic diversity of crops, the wide variety of pests and diseases. As well as the characteristics of the data sets, the types, and peculiarities of network architectures and models convolutional neural and the complexity involved in the results optimization techniques.

Thus, this systematic literature review (SLR) is intended to characterize the state of the art recognition and identification of plant diseases using convolutional neural networks. The investigation sought significant contributions regarding the challenges listed and the different innovations that aim to improve the performance of CNNs and, consequently the correct identification of diseases. A systematic literature review is a type of scientific research that aims to present an unbiased report about a research topic, following a methodology that is reliable, accurate and that allows auditing~\cite{kitchenham:2004,Fabbri:2013}.

The remainder of the paper is organized as follows: In Section~\ref{S:2}  presented the background. Section~\ref{S:3} describes our review protocol in detail. Section~\ref{S:4} presents the results of the SLR. Section~\ref{S:5} presents the discussion and finally Section~\ref{S:6} concludes the paper.

\section{Background}
\label{S:2}

The identification of plant diseases is one of the most basic and essential agricultural activities. In most cases, identification is performed manually, visually, by serological and molecular tests or by microscopy. The problem with the visual assessment to identify diseases is that the evaluating subject has assumed a subjective task, prone to psychological and cognitive phenomena that can lead to prejudice, optical illusions, and, finally, error\cite{BARBEDO:2016}.

\subsection{Types of Plant Diseases}
Plant diseases are classified according to the type of their causative agent. Diseases originating from living organisms are called biotic, unlike diseases that are produced by non-living ecological circumstances, called abiotic\cite{Husin:2012}. Fungi, bacteria, viruses, plague, nematodes are the leading causes of
different forms of biotic diseases\cite{Agrios:2005}. In this study, the pest or plague, or more specifically, biological pest, is the outbreak of certain species harmful to agricultural development that cause epidemic diseases in plants and their cultivars.

\subsection{Plant Disease Detection System}

The automated solutions for the identification of plant diseases using images and machine learning, especially CNNs, have provided significant advances to maximize the accuracy of correct diagnosis. Convolutional Neural Networks is a subset of machine learning approaches that have emerged as a versatile tool for assimilating large amounts of heterogeneous data and providing reliable predictions of complex and uncertain phenomena\cite{Goodfellow:2016}.

\subsection{Evolution of CNN architectures}

Different innovations in CNN architectures have been proposed since 1998 with the presentation of LeNet-5\cite{LeNet:1998}. It can be categorized as parameter optimization, regularization, structural reformulation, etc. However, it is observed that the main improvement in the performance of CNNs is motivated by the fundamental reformulation and design of new blocks\cite{ArchictetureCNN:2019}. Thus, this SLR classifies as new architecture that is by the innovation proposals reported in studies of \cite{ArchictetureCNN:2019} and \cite{Goodfellow:2016}.

\subsection{Transfer Learning}

Transfer learning is a method that reuses models applied to specific tasks as a starting point for a model related to a new domain of interest. Thus, the objective is to borrow labeled or extract knowledge from some related fields to obtain the highest possible performance in the area of interest\cite{TransferLearning:2010,TransferLearning:2015}.

As per standard practices, there are two ways to approach transfer learning \cite{Khandelwal:2019}:

\begin{itemize}
    \item \textbf{Using the base neural network as a fixed feature extractor:} In this case, the images of the target dataset are fed to the neural network, and the features that generate as input to the final classifier layer are extracted. Through these features, a new classifier is built, and the model is created.

    \item \textbf{Fine Tuning the base network:} In this case, the final classifier layer is replaced, just like the above case but using backpropagation algorithms, the weights of previous layers are also modified.
\end{itemize}

\subsection{Custom Layers}

Traditional CNN Architectures build their models using three types of layers, namely Input, Intermediate, and Output\cite{ArchictetureCNN:2019}. Generally, Frameworks have a list of standardized layers according to the most relevant models that already exist. However, when the peculiarities of the problem demand a need to operate at a lower level of abstraction, the need arises to implement a customized or personalized layer that meets the specific demand. For example, when using ResNet for a particular purpose, there is a need to add an extra layer to the end of the model that normalizes the classifier output.

In this way, Custom Layers are defined as those that are not included in a list of known layers and implemented natively by the Frameworks, through the most relevant models. It should be noted that the implementation of customized layers based on an existing architecture or model, does not constitute a new architecture or model\cite{CustomLayer:2019}.

\subsection{Data Augmentation}

One of the successful approaches that seek to mitigate the problem of the ability to generalize models is data augmentation. In this approach, the original data set is augmented by several techniques seeking a more comprehensive representation possible of the problem domain, minimizing the distance between the training and validation set \cite{Shorten:2019}.

The data augmentation uses the methods of \textit{(Data Warping)} or super-sampling \textit{(Oversampling)}, that artificially increase the size of the training data set. Data distortion transforms existing images so that their label is preserved, using techniques such as geometric transformations, color space, image slices, rotation, translation, adding noise, kernel filters.



\section{Planning Review}\label{S:3}

The planning of this SLR followed the processes defined by \cite{Fabbri:2013}. To provide transparency and allow the replication and auditing of the entire process (essential requirements in systematic reviews), the StArt Tool (State of the Art Through Systematic Review) was used \cite{Start:2012}, which supports all phases of an SLR.

In the planning phase, we need to identify the real need, namely the motivation for implementing the SLR. Thus, defining the objectives and elaborating a protocol for SLR are essential elements for successful execution of the technique. It should be noted that the quality of the protocol directly impacts the SLR results.

\subsection{Research Questions}

It is known that CNNs have been showing better results than traditional Computer Vision techniques and other techniques using Machine Learning. Thus, this RSL is intended to identify the characteristics that affect the efficiency of plant disease identification approaches using convolutional neural networks. And the main research question of SRL that we intend to answer is: 

\begin{center}
\textit{"How are convolutional neural networks innovating and overcoming the challenges of the plant disease identification process?"
}
\end{center}

Additionally, secondary questions were formulated to assist in the identification of characteristics that determine the desired response, namely:

 \begin{itemize}
 \footnotesize{
     \item[SQ1]$-$ Which approaches make use of new architectures or models?
     
     \item[SQ2]$-$ What are the characteristics of the data sets predominantly used?

     \item[SQ3]$-$ What types of crops are most investigated with approaches using CNNs?
     
     \item[SQ4]$-$ What types of approaches and frameworks are commonly used?
     
     \item[SQ5]$-$ Which CNN algorithm is prevalent in current approaches?
     
     \item[SQ6]$-$ What types of plant diseases are most investigated with approaches using CNNs?
     
}
\end{itemize}

\subsection{Search Strategy}

This SLR focuses its research on digital scientific databases, so any type of gray literature was not included in the investigation process. The study is based on the assumption that most of the research with relevant results that may appear in the gray literature are typically described or referenced in scientific papers already published.

The search and selection strategy for primary studies focused on 5  indexed electronic databases, namely: ACM Library~\footnote{https://dl.acm.org access:January/2020}, IEEE Xplore Digital Libray~\footnote{https://ieeexplore.ieee.org/Xplore access:January/2020}, Elsevier Scopus~\footnote{https://www.scopus.com access:January/2020}, Springer Link~\footnote{https://link.springer.com access:January/2020} and Google Academic~\footnote{https://www.scholar.google.com access:January/2020}. The databases were selected because they offer the largest volume of high-impact scientific conference proceedings and journals, broadly covering the field of investigations on plant diseases and machine learning through CNNs.

According to \cite{Petersen:2008} and \cite{Fabbri:2013}, the process of collecting the study set begins, by defining the \textit{Search String}. Thus, the \textit{Search String} should be formulated based on the experience of the researchers and reviewers involved in the SLR process. Even though different indexed research repositories define their syntax for building search strings, we initially defined the following generic string that was later customized for each search engine:
\begin{center}
\footnotesize
\textit{
(``plant disease" OR  ``plant pathology" OR ``crop disease")\\
AND\\
(``machine learning" OR ``convolutional neural network" OR ``deep learning" OR ``DNN" OR ``CNN")}

\end{center}

\subsection{Study Selection Criteria}

The identification and recognition of plant diseases through convolutional neural networks has aroused a high interest in the scientific community in the last decade. Therefore, publications from the previous ten years (from 2010 to 2019) were selected.

The search string had a broad scope intentionally because we did not want to miss any potentially exciting research. This full scope has led to a large number of papers, from which we filtered the most
relevant ones using the selection criteria presented in Table~\ref{tab:SelectionCriteria}.

\begin{table}[!ht]
\centering
\caption{Selection criteria for studies defined in the SLR planning protocol}
\label{tab:SelectionCriteria}
\resizebox{\textwidth}{!}{%
\begin{tabular}{cccl}
\hline
\multicolumn{1}{l}{\multirow{2}{*}{ID}} & \multicolumn{2}{c}{Creterion Type} &\multicolumn{1}{c}{\multirow{2}{*}{Description}} \\ \cline{2-3}
\multicolumn{1}{l}{} & \multicolumn{1}{l}{Inclusion} \vline& \multicolumn{1}{l}{Exclusion} &  \\ \hline
C1 & X & - & Studies that approaches the identification of plant diseases through convolutional neural networks. \\ 
C2 & X & - & Studies that present new models or architectures of convolutional networks. \\ 
C3 & X & - & Studies describing techniques or methods for identifying plant diseases using CNNs. \\ 
C4 & - & X & Studies without full text available \\ 
C5 & - & X & Studies not written in English \\ 
C6 & - & X & Duplicate publication from multiple sources \\ 
C7 & - & X & Studies without qualifying data for extraction \\ 
C8 & - & X & Studies that do not use CNNs as the main approach \\ \hline
\end{tabular}%
}%
\end{table}

Once identified, studies need to be selected through the application of selection criteria (which are the inclusion and exclusion criteria) and can be evaluated by the quality criteria. The selection criteria must specify the main characteristics and/or content that the studies must have to be included or excluded.

\subsection{Study Quality Assessment}

The quality criteria aim to evaluate methodological aspects of the studies, that is, they can be considered issues such as the relevance of the research theme and the use of methods that lead to the objectives proposed in the study.

The quality assessment is performed using forms developed by the researchers. These are composed of quality criteria, that is, questions about methodological aspects of each study\cite{Kitchenham:2012, Staples:2007}. Table~\ref{tab:QualityCriteria} presents the items that make up the study quality assessment form.

\begin{table}[!ht]
\centering
\caption{Criteria for Quality Assessment of studies defined in the SLR planning protocol.}
\label{tab:QualityCriteria}
\resizebox{\textwidth}{!}{%
\begin{tabular}{llllll}
\hline
\multicolumn{1}{c}{\multirow{2}{*}{ID}} & \multicolumn{1}{c}{\multirow{2}{*}{Question}} & \multicolumn{3}{c}{Indicator} & \multicolumn{1}{c}{\multirow{2}{*}{Reference}} \\ \cline{3-5}
\multicolumn{1}{c}{} & \multicolumn{1}{c}{} & \multicolumn{1}{c}{\textbf{Y}} & \multicolumn{1}{c}{\textbf{P}} & \multicolumn{1}{c}{\textbf{N}} & \multicolumn{1}{c}{} \\ \hline
\multirow{3}{*}{Q1} & \multirow{3}{*}{Are the studies cited by other works and scientific investigations?} & \multicolumn{1}{c}{\multirow{3}{*}{1,0}} & \multicolumn{1}{c}{\multirow{3}{*}{0,5}} & \multicolumn{1}{c}{\multirow{3}{*}{0,0}} & \textbf{Yes} $\Rightarrow$ Studies published studies with a significant number of citations. \\ \cline{6-6} 
 &  & \multicolumn{1}{c}{} & \multicolumn{1}{c}{} & \multicolumn{1}{c}{} & \textbf{Partially} $\Rightarrow$ Studies published with a low number of citations \\ \cline{6-6} 
 &  & \multicolumn{1}{c}{} & \multicolumn{1}{c}{} & \multicolumn{1}{c}{} & \textbf{Not} $\Rightarrow$ Studies published without citations \\ \hline
\multirow{3}{*}{Q2} & \multirow{3}{*}{\begin{tabular}[c]{@{}l@{}}Do the studies describe the approaches in detail and provide subsidies \\ for replication and finding results?\end{tabular}} & \multicolumn{1}{c}{\multirow{3}{*}{1,0}} & \multicolumn{1}{c}{\multirow{3}{*}{0,5}} & \multicolumn{1}{c}{\multirow{3}{*}{0,0}} & \textbf{Yes} $\Rightarrow$ Studies explicitly describe the approaches; \\ \cline{6-6} 
 &  & \multicolumn{1}{c}{} & \multicolumn{1}{c}{} & \multicolumn{1}{c}{} & \begin{tabular}[c]{@{}l@{}}\textbf{Partially} $\Rightarrow$ Studies describe the approaches, however, they do not provide \\ subsidies for replication and verification of results\end{tabular} \\ \cline{6-6} 
 &  & \multicolumn{1}{c}{} & \multicolumn{1}{c}{} & \multicolumn{1}{c}{} & \begin{tabular}[c]{@{}l@{}}\textbf{Not} $\Rightarrow$ Studies that do not provide sufficient characteristics to describe \\ the approaches, nor do they allow replication and auditing;\end{tabular} \\ \hline
\multirow{3}{*}{Q3} & \multirow{3}{*}{\begin{tabular}[c]{@{}l@{}}Are the main findings stated clearly? regarding creditability, validity, and\\ reliability?\end{tabular}} & \multirow{3}{*}{1,0} & \multirow{3}{*}{0,5} & \multirow{3}{*}{0,0} & \textbf{Yes} $\Rightarrow$ The findings are clearly reported \\ \cline{6-6} 
 &  &  &  &  & \textbf{Partially} $\Rightarrow$ The main findings are partially reliable and credible. \\ \cline{6-6} 
 &  &  &  &  & \textbf{Not} $\Rightarrow$ There is no credibility in the findings presented. \\ \hline
\multirow{3}{*}{Q4} & \multirow{3}{*}{Are the limitations of the studies clearly specified and documented?} & \multirow{3}{*}{1,0} & \multirow{3}{*}{0,5} & \multirow{3}{*}{0,0} & \textbf{Yes} $\Rightarrow$ Specifies and documents study limitations \\ \cline{6-6} 
 &  &  &  &  & \begin{tabular}[c]{@{}l@{}}\textbf{Partially} $\Rightarrow$ It mentions the limitations, however it does not specify \\ or document its determinants.\end{tabular} \\ \cline{6-6} 
 &  &  &  &  & \begin{tabular}[c]{@{}l@{}}\textbf{Not} $\Rightarrow$ It does not mention the limitations and threats of validity \\ of the studies.\end{tabular} \\ \hline
\multirow{3}{*}{Q5} & \multirow{3}{*}{Are the conclusions credible and consistent with the results presented?} & \multirow{3}{*}{1,0} & \multirow{3}{*}{0,5} & \multirow{3}{*}{0,0} & \begin{tabular}[c]{@{}l@{}}\textbf{Yes} $\Rightarrow$ Methodologically prepared studies with consistency of facts and \\ confidence in the discoveries presented\end{tabular} \\ \cline{6-6} 
 &  &  &  &  & \begin{tabular}[c]{@{}l@{}}\textbf{Partially} $\Rightarrow$ Methodologically elaborated studies, but with insufficient \\ details to certify the results and conclusions.\end{tabular} \\ \cline{6-6} 
 &  &  &  &  & \textbf{Not} $\Rightarrow$ Studies methodologically poorly elaborated and without credibility. \\  \cline{1-6}
\end{tabular}%
}

\end{table}

We chose the criteria based on their influence on the quality of
the final product. Points were given to each of the five criteria based on the following scale: yes = 1, partially = 0.5, not = 0.

The quality of each study is directly associated with its methodological quality, which can be measured by aspects such as internal validity (effectively measuring what you want to measure), external validity (ability to generalize the results), the relevance of the research topic (questions research and well-defined objectives and based on the literature), adoption of methods that lead to the proposed goals, among others.

Primary studies that obtained a quality index (IQ) of less than 3 points were excluded from this SLR, to guarantee the high quality of the summarized final set.

\subsection{Data Extraction and Synthesis}

To elect the primary studies, we use the PRISMA~\textit{(Preferred
Reporting Items for Systematic and Meta-Analyses}) recommendation, according to its four stages: identification, screening, extraction, and final selection\cite{PRISMA:2009}. In the identification, the study sample was extracted, quantified, and stored eliminating duplicate studies. In the screening stage, the reviews were pre-selected from reading the title, abstract, and keywords by two independent reviewers, obtaining a Kappa index $>$ 80\% ($p <0.05$) of interobserver agreement. Discrepancies were resolved by a third reviewer\cite{Buscemi:2006}.

In the extraction stage, the full reading of pre-selected studies was performed, excluding studies that did not meet the inclusion criteria, defining the final sample. The synthesis activity aims to combine the data extracted from each of the selected primary studies, delineating the situation of the investigated research questions.

A total of 2,221 studies were retrieved, where 1,433 were identified in the Google Academic, 322 in the IEEE Library, 233 in the ACM Library database, 195 in the Elsevier Scopus, and 38 in the Springer Link.

\begin{figure}[!ht]
\begin{center}
\includegraphics[width=10cm]{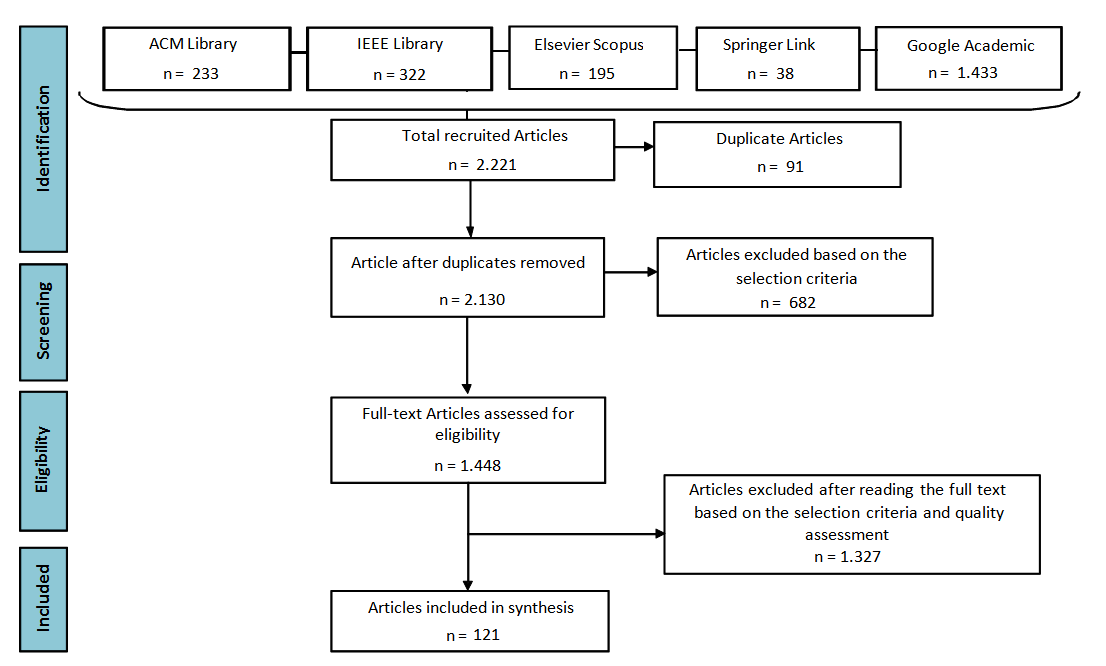}
\end{center}
\vspace{-.3cm}
\caption{Flowchart of study phases and selections - adapted from \cite{PRISMA:2009}}
\label{Fig:fluxograma_sumarizacao}
\end{figure}

During the identification phase, 91 duplicate studies were excluded, and during the initial selection, 682 reviews were eliminated because they did not meet the established inclusion criteria. For the extraction phase, 1,448 reviews were selected, and, after reading the entire text, 121 reviews were summarized. Figure~\ref{Fig:fluxograma_sumarizacao} illustrates the steps followed in conducting the SLR with a summary of the results.

\begin{figure}[!ht]
\begin{center}
\includegraphics[width=10cm]{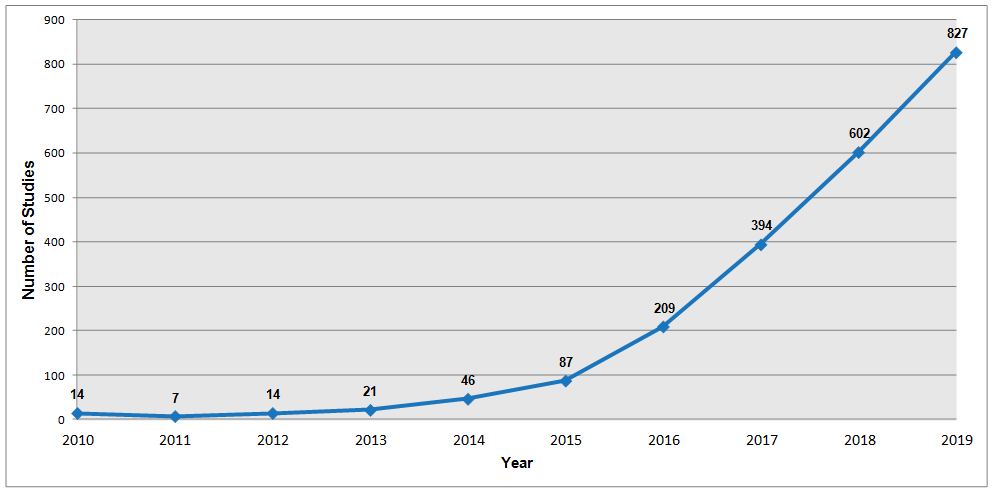}
\end{center}
\vspace{-.3cm}
\caption{Temporal evolution of the quantity of publications recruited containing the keywords defined in the Search String in the last ten years (2010 to 2019)}
\label{Fig:grafico_linhasSelecao}
\end{figure}

The investigation was carried out retrieving studies between the years 2010 to 2019, it is possible to note that the year 2012 is a milestone not only for studies that address convolutional networks, but also the application of this approach to the identification and recognition of plant diseases. Figure \ref{Fig:grafico_linhasSelecao} presents the evolution of studies that address this theme.

\begin{figure}[!ht]
\begin{center}
\includegraphics[width=13cm]{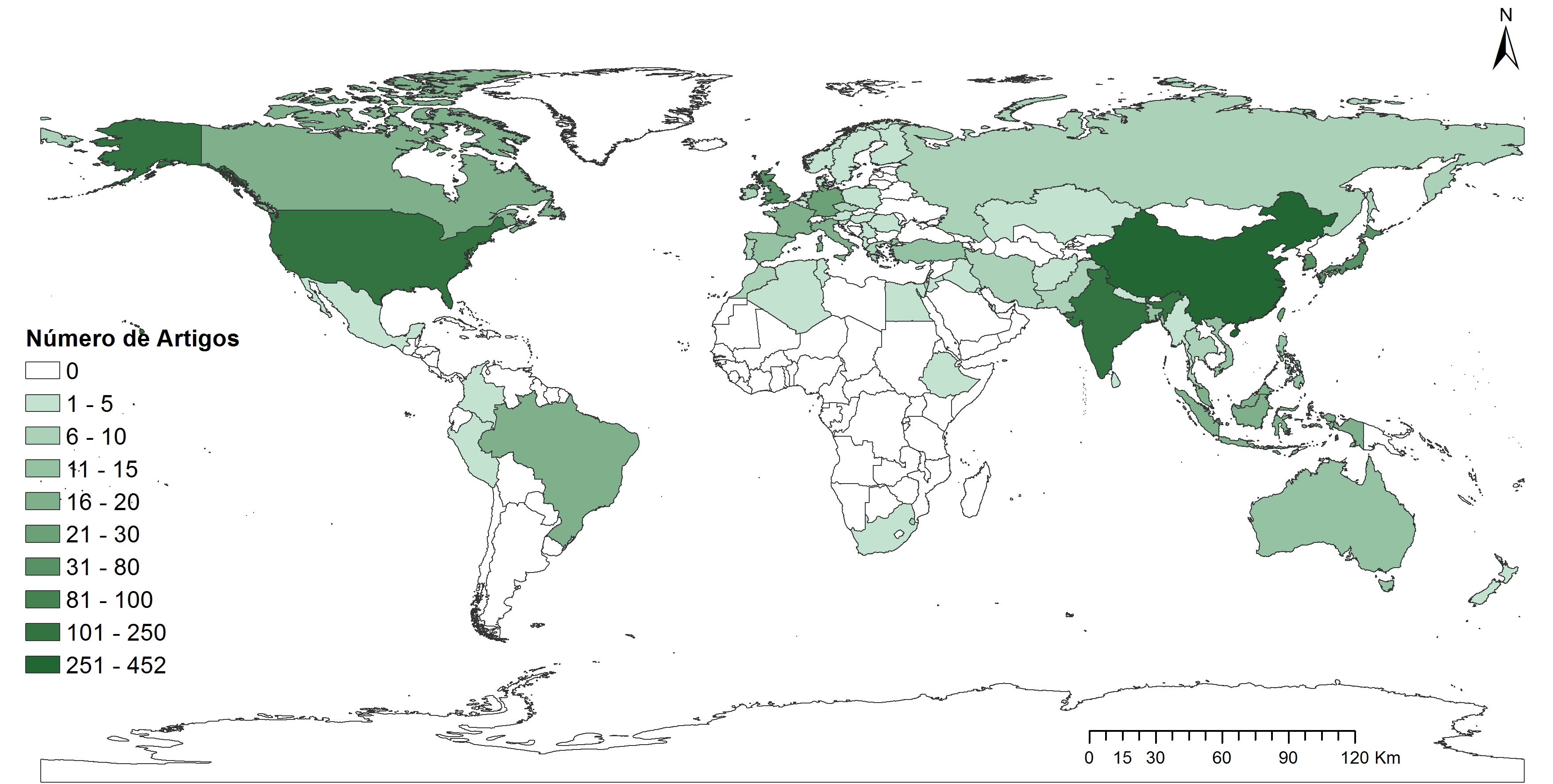}
\end{center}
\vspace{-.3cm}
\caption{Geographical distribution of the studies identified in the SRL}
\label{Fig:mapa_selecionados}
\end{figure}

Additionally, the studies identified were distributed geographically, as illustrated by Figure \ref{Fig:mapa_selecionados}. It was noted that the investigations and consequently the scientific interest on the identification and recognition of plant diseases using machine learning techniques are coincident with the most populous countries in the world, among them: China, India, the United States of America, Indonesia, Pakistan, and Brazil.

At the end of the synthesis activities, Table \ref{Tab:finalSelect} was prepared, available in Appendix A with 121 primary studies chosen for extraction. Basically, the table is composed by the following attributes: ID (Study identification), Refs (Author of the approach), IQ (Quality Index), Algorithm, Best accuracy, Dataset characteristics, Crop type, Image type, Language and Framework , Predominant approach. In addition, in Table \ref{Tab:PathogenStudies}, information about each investigated disease was detailed according to the type of crops and causative pathogen. Information from Table \ref{Tab:PathogenStudies} is available in~\ref{appendiceB}.

Each column of the Table summarizes a relevant feature of the investigation process that is directly related to the questions formulated for the SLR.

\section{Results}
\label{S:4}

In this section, we present the discrete and continuous quantitative data extracted from the 121 primary studies that support formulating the answers corresponding to the six secondary questions of the research.

\subsection{SQ1: Which approaches make use of new architectures or models?} \label{SQ1}


To answer this question, the data extraction from the 121 summarized studies made it possible to group the investigations by practical approaches. Seven groups of methods were categorized, namely: Customizing Layers, Transfer Learning, New Architecture, Fine Tune and hyperparameters, Image Segmentation, Data Augmentation, and Unsupervised Learning.

\begin{figure}[!ht]
\begin{center}
\includegraphics[width=13cm]{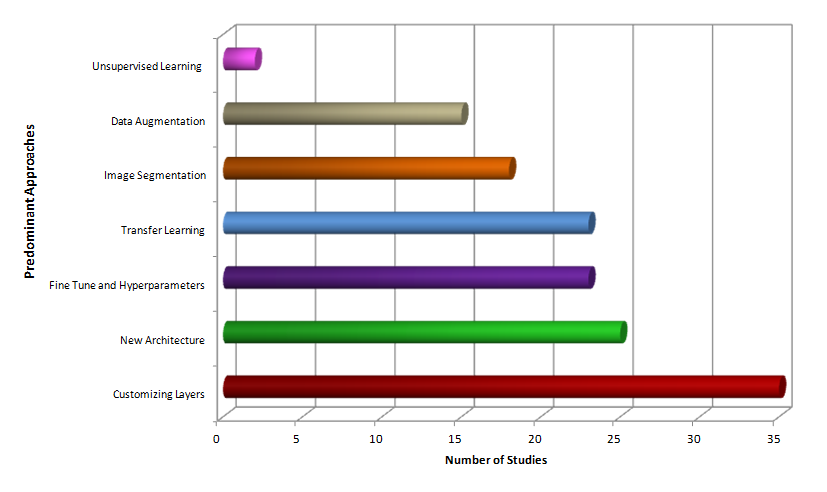}
\end{center}
\vspace{-.3cm}
\caption{Predominant Approaches in 121 Summary Studies.}
\label{Fig:PredominantApproaches}
\end{figure}

Among the 121 studies evaluated, it is possible to notice that only 25 studies (S1, S2, S6, S7, S11, S12, S13, S15, S22, S35, S43, S54, S61, S65, S68, S70, S75, S84, S86, S91, S100, S113, S114, S116, S120) propose a new architecture. In Figure \ref{Fig:PredominantApproaches}, the distribution of studies by practical approaches is presented.

\subsection{SQ2: What are the characteristics of the data sets predominantly used?} \label{SQ2}

\begin{figure}[!ht]
\begin{center}
\includegraphics[width=13cm]{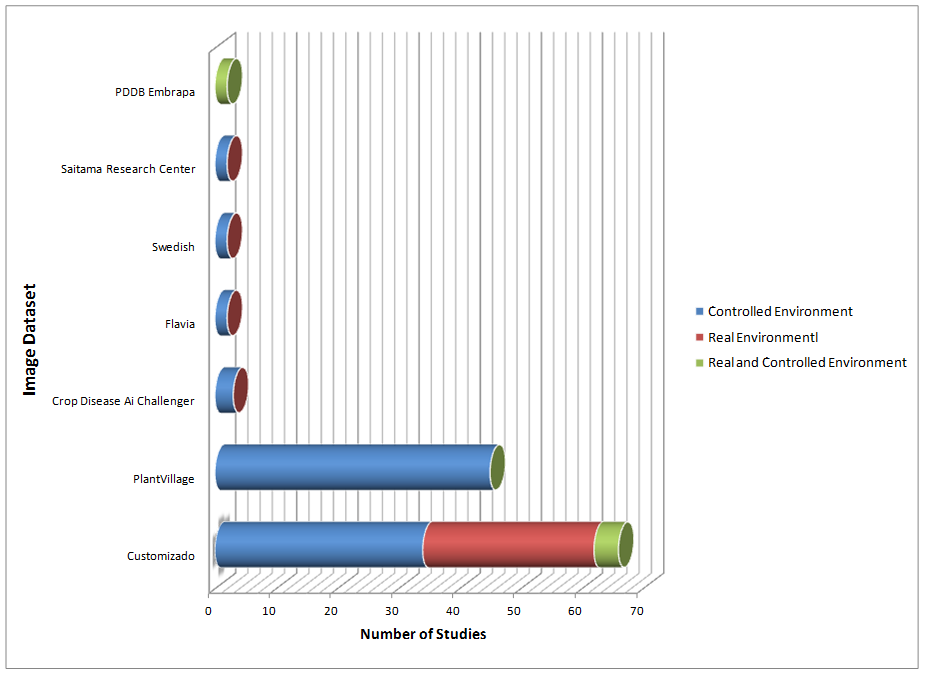}
\end{center}
\vspace{-.3cm}
\caption{Characteristics of the Data Sets present in the 121 Summary Studies}
\label{Fig:DatasetRSL}
\end{figure}

Regarding the predominance and characteristics of the data sets most used in the investigated studies, we highlight the customized initiatives designed mainly for a research need and the PlantVillage\cite{PlantVillage:2016} data set. Figure~\ref{Fig:DatasetRSL} shows the distribution of studies by data set and type of image capture environment.

The customized Data Sets are present in 66 studies (S1, S2, S3, S6, S7, S8, S9, S12, S15, S17, S18, S20, S22, S24, S25, S28, S32, S33, S34, S36 , S37, S38, S40, S43, S44, S45, S46, S47, S49, S50, S52, S58, S64, S68, S69, S70, S71, S73, S74, S75, S77, S79, S80, S82, S84 , S87, S88, S89, S90, S91, S92, S95, S96, S97, S98, S100, S102, S103, S104, S107, S112, S113, S115, S116, S119, S121) and PlantVillage is used in your full or partial version in 45 studies (S3, S4, S9, S11, S13, S14, S18, S21, S23, S26, S27, S28, S29, S35, S41, S48, S51, S53, S55, S56, S57, S60, S61, S62, S63, S65, S66, S67, S68, S69, S72, S73, S76, S78, S81, S85, S86, S93, S99, S101, S105, S108, S110, S114, S117).

\subsection{SQ3: What types of crops are most investigated with approaches using CNNs?}

\begin{figure}[!ht]
\begin{center}
\includegraphics[width=13cm]{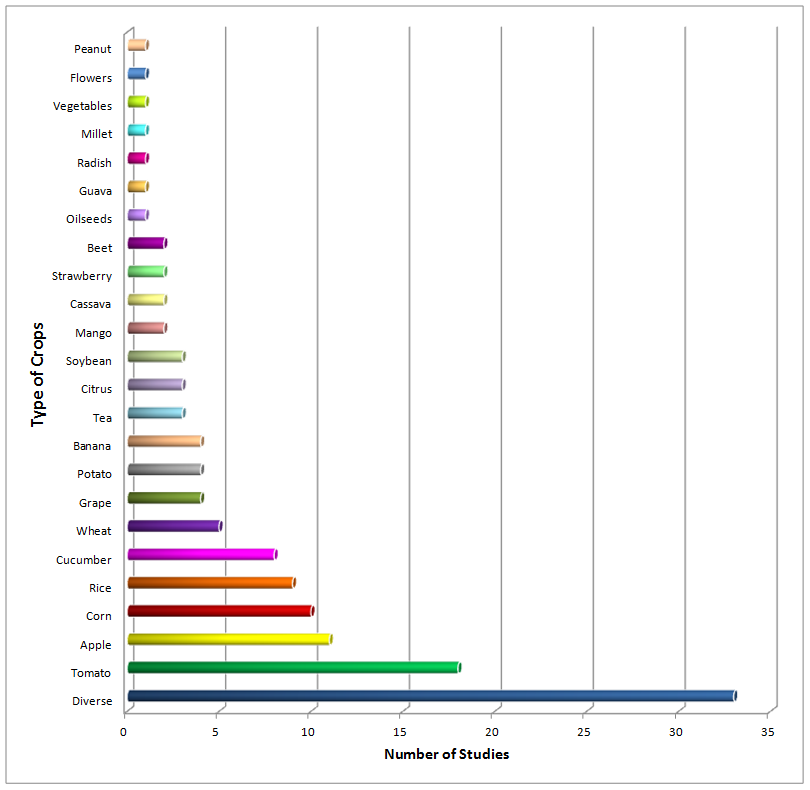}
\end{center}
\vspace{-.3cm}
\caption{The most investigated crops in the 121 studies that apply CNNs to identify diseases}
\label{Fig:TypeCrops}
\end{figure}

The types of crops present in the approaches that use CNNs to identify diseases are summarized graphically and presented in Figure~\ref{Fig:TypeCrops}. It is possible to observe that the type called "Diverse" categorizes all studies that use more than one type of crop in their approaches and consequently make use of a database with diversified plant phenotypes. In particular, this is a predominant feature in the PlantVillage data set.

Studies that investigate only one type of crop in particular highlight those that address the phytopathology of Tomato crop with 18 investigations (S3, S9, S27, S44, S48, S55, S58, S61, S67, S70, S75, S78, S86, S93, S99, S102, S105, S108).

\subsection{SQ4: What types of approaches and frameworks are commonly used?}

\begin{figure}[!ht]
\begin{center}
\includegraphics[width=13cm]{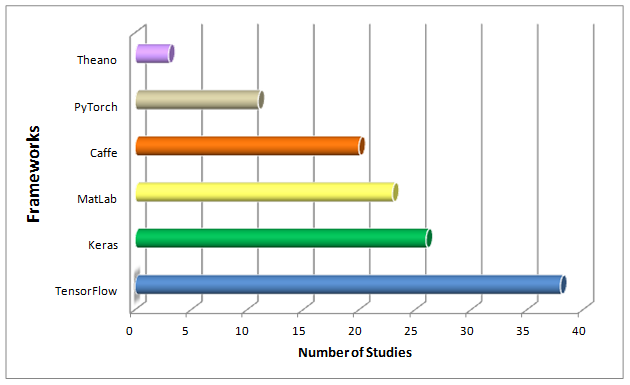}
\end{center}
\vspace{-.3cm}
\caption{The most used frameworks among the 121 studies summarized}
\label{Fig:FrameworkRLS}
\end{figure}

In the summary process, it was found that the investigated studies make use of 6 main FrameWorks, used to implement solutions involving machine learning, as well as predictive analysis. Figure~\ref{Fig:FrameworkRLS} is apresented  the distribution of studies grouped by the framework used in the approach development process.

TensorFlow is the most used framework in approaches that investigate plant diseases through convolutional neural networks. Among the 121 summarized studies, 38 investigations (S1, S2, S3, S5, S8, S11, S12, S14, S15, S23, S25, S27, S28, S38, S39, S40, S42, S44, S50, S53, S55 , S56, S61, S62, S63, S64, S65, S66, S68, S69, S70, S76, S77, S81, S86, S104, S106, S110) use the TensorFlow.

\subsection{SQ5: Which CNN algorithm is prevalent in current approaches?}

\begin{figure}[!ht]
\begin{center}
\includegraphics[width=13cm]{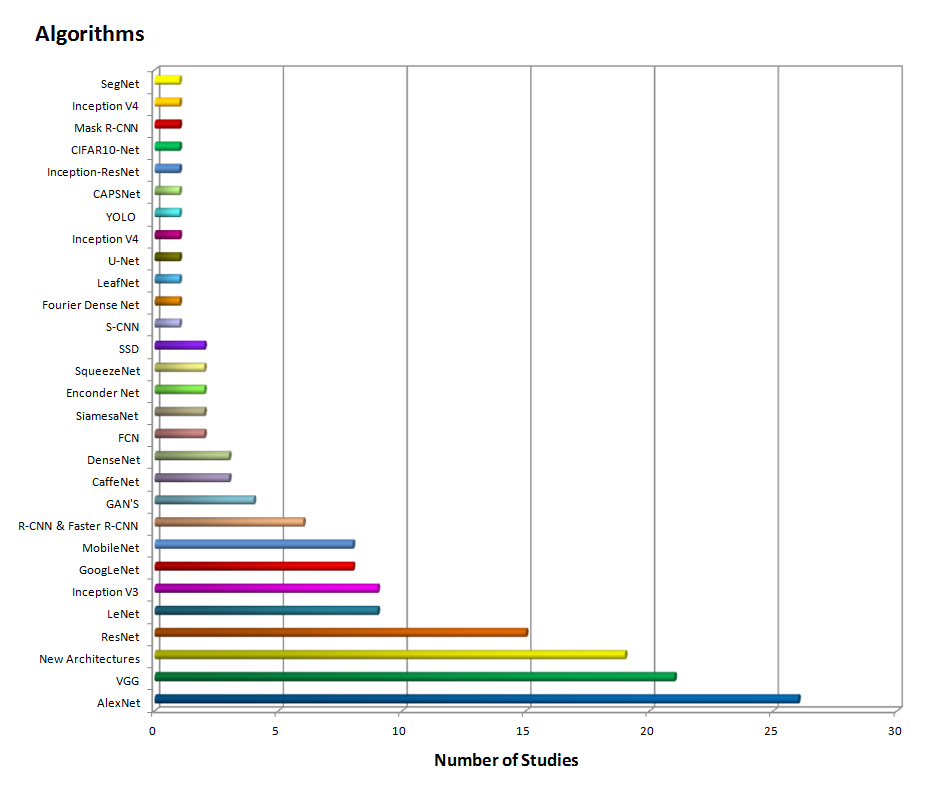}
\end{center}
\vspace{-.3cm}
\caption{The CNN Algorithms most used in the 121 summarized studies}
\label{Fig:Algorithms}
\end{figure}

Deep learning is not a single approach, but a class of algorithms and topologies that can be applied to a wide variety of problems. In Figure~\ref{Fig:Algorithms}, it is possible to follow the distribution of these algorithms grouped according to the 121 summarized studies.

Thirty algorithms used in the approaches were accounted for, with predominant algorithms that define the classic architectures, with emphasis on AlexNet and VGG. It should be noted that 19 studies (S1, S6, S7, S11, S12, S13, S22, S35, S43, S61, S68, S70, S75, S84, S86, S91, S100, S116, S120) propose new architectures and 15 (S9, S18, S19, S23, S32, S34, S39, S41, S42, S49, S50, S51, S53, S65, S103) investigations customize classic algorithms to solve the problem of identifying plant diseases using CNNs.

\subsection{SQ6: What types of plant diseases (biotic causative agents) are most investigated with approaches using CNNs?}

\begin{figure}[!ht]
\begin{center}
\includegraphics[width=13cm]{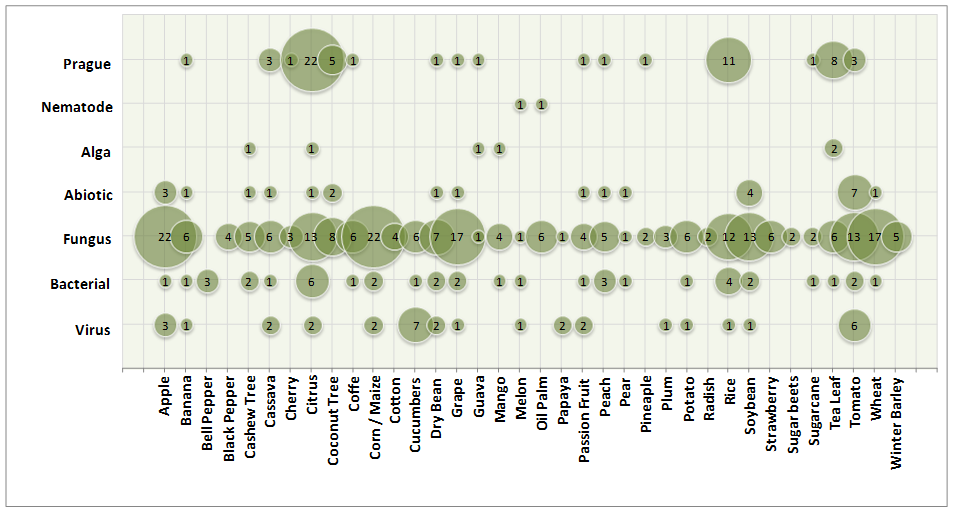}
\end{center}
\vspace{-.3cm}
\caption{Relationship of type of crops X groups of pathogens X occurrences of diseases investigated by causative agents}
\label{Fig:Biotic_Abiotic}
\end{figure}

Using the data summarized in~\ref{appendiceB}, we analyzed 54 datasets with 142 types of crops,  associating each disease-causing pathogen to its large group, namely: Virus, Bacteria, Fungi, Algae, Plague, Nematodes and, Abiotics. In Figure \ref{Fig:Biotic_Abiotic}, we present the type relationship of the crops with groups of pathogens, quantifying the number of occurrences of diseases investigated by their causative agents.

\section{Discussion}
\label{S:5}

To keep the discussion in line with the core questions, all of them answered in the previous section, our study is organized into four topics that describe state of the art on identifying plant diseases using CNNs.

\subsection{Predominant Approaches} \label{S:51}

Different innovations in CNN architectures have been proposed since 1998 with the presentation of LeNet-5\cite{LeNet:1998} and can be categorized as parameter optimization, regularization, structural reformulation (Convolutional layer, Polling layer, and Activation Function), loss function and fast processing. However, it is observed that the main improvement in the performance of CNNs is motivated by the structural reformulation and design of new blocks\cite{ArchictetureCNN:2019}.

When we tried to answer the question SQ1 of Section~\ref{SQ1}, we noticed that the investigations that applied state of the art for identification and classification of plant diseases using CNNs explored different innovative approaches to provide a reliable solution to the problem investigated.

As for the new architectures, we noticed three trends for the formulation of approaches. One of them is the structural reformulation of traditional architectures, such as LeNet (studies S61, S65, S68, S84, S100, S114, and S116), AlexNet (S113 and S120) and ResNet (S6, S43, S75, and S86). Among these, it is possible to highlight the S84 approach proposed by \cite{Ghosal:2018} that uses LeNet as an explainable model that can identify and quantify leaf tensions consistently, quickly, and accurately. The proposal seeks a solution for mobile platforms for detection in real-time with an accuracy of 95.04\%, aiming at a fast and large-scale observation of crops in real cultivation environments.

The S113 study proposed by \cite{LiuZhang:2017} describes an approach for the classification and identification of four common diseases of apple leaves. The architecture achieves an overall accuracy of 97.62\% and a significant reduction in the number of parameters compared to a standard AlexNet model. Another architecture based on ResNet, called HOResNet, is proposed in the study S86 de \cite{Zeng:2018} aiming at a robust recognition of plant diseases. The research explores the problem of low precision in the identification and classification of plant diseases with images acquired in real crop environments. It improves the anti-interference ability by exploring images with sizes, shooting angles, poses, background, and lighting in different aspects. The results demonstrate that the approach reaches an accuracy of 91.79\% and 90.14\% to diseases of Tomato and Rice and Cucumbers, respectively.

Another trend is the blending of blocks between architectures, such as the S15 study (Yolo-DenseNet) and the S22 study (Inception-ResNet). In the \cite{Padilla:2019} proposal of study S15, a CNN-based method of detecting anthracnose lesions is described using data augmentation techniques using Cycle-Consistent Adversarial Network (CycleGAN). This approach includes a DenseNet to optimize the resource layers of the YOLO-V3 model that have low resolution, reaching an accuracy of 95.57\%. The S22 study by \cite{XinZhang:2019} proposes a new deep convolutional neural network (DCNN) based approach for auTomatod crop disease detection using very high spatial resolution hyperspectral images captured with UAVs. The proposed model introduced multiple Inception-ResNet layers for feature extraction and was optimized to establish the most suitable depth and width of the network. The proposal deals with three-dimensional data, using spatial and spectral information to detect yellow rust in wheat with an accuracy of 85\%.

The latest trend includes proposals that implement architectures explicitly designed for the identification and classification of plant diseases, with emphasis on models that use hyperspectral images, namely: GPDCNN (S1), 3D-CNN (S7), PDDNN (S11), M-bCNN-CKM (S12), Teacher / Student Architecture (S13), PlantDiseaseNet (S35), Fourier Dense Network (S54) and OR-AC-GAN (S70).

In the S70 study proposed by \cite{Wangx:2018}, an approach called OR-AC-GAN is described, developed for the early detection of Tomato spotted wilt virus using hyperspectral images and an auxiliary external removal classifier using opposing generation networks. The proposal integrates the tasks of plant segmentation, spectrum classification, and image classification. The results showed that the accuracy reaches 96.25\% before the visible symptoms appear. The approach proposed by \cite{Nagasubramanian:2019} in study S7 projects a CNN architecture called supervised 3D-CNN based model to learn the spectral and spatial information of hyperspectral images for classification of healthy and charcoal rot infected samples. A saliency map-based visualization method is used to identify the hyperspectral wavelengths that make a significant contribution to classification accuracy. The model achieved a classification accuracy of 95.73\%.




\subsection{Characteristic of the Data Sets} \label{S:52}

The characteristics of the data sets, such as the number of samples, number of diseases, symptomatic similarity caused by pathogens, pests, and abiotic factors, constitute one of the most important limitations regarding the accuracy of the approaches that implement the identification and classification of diseases of plants. Experimental results indicate that while the technical constraints linked to automatic plant disease classification have mainly been overcome, the use of limited image datasets for training brings many undesirable consequences that still prevent the effective dissemination of this type of technology \cite{BARBEDO:2018}.

The answer to question SQ2 from Section \ref{SQ2}, groups the data sets with the highest recurrence in the studies, and demonstrates that the approaches try to reduce the limitations by creating customized sets that mix images from controlled environments with images captured under real growing conditions. However, we noticed that 65.28\% of the approaches use sets with images in controlled environments, with PlantVillage being present in 37.19\% of the summarized studies.

It should be noted that many approaches, such as studies S28, S35, S40, S65, S69, and S73, have already demonstrated a reduction in accuracy, when they use models trained exclusively with sets of images acquired in controlled environments to classify images in challenging conditions such as illumination, complex background, different resolution, size, pose, and orientation of real scene images. According to \cite{Goncharov:2019}, who proposed the approach of the S28 study, when applying a customized CNN architecture, they obtained high precision with training and testing using the PlantVillage data set. However, the results obtained when applying the same network in images acquired in real crop conditions, there was a reduction of about 30\% to 40\% in the global accuracy.

\subsection{CNN Algorithms and Frameworks} \label{S:53}

Among the summarized studies, TensorFlow\cite{Tensorflow:2015} is the framework for creating and training CNN networks most used by approaches. Following this trend, Keras\cite{Keras:2015} is a high-level neural networks API, written in Python, and capable of running on top of TensorFlow, which demonstrates significant results regarding the implementation of approaches using CNNs.

As for the implementation of these CNN architectures, we note that traditional algorithms such as, for example, AlexNet, VGG, ResNet, LeNet, InceptionV3, GoogLeNet are predominant in the investigated studies. The recurring use of these implementations is explained by their success cases based on the ImageNet challenge \cite{ILSVRC:2014}. Approximately 85\% of the approaches use the transfer learning, fine tune, and hyperparameters methods to increase the accuracy of their results in detriment to the deficiencies of the data sets used. Additionally, it is possible to relate the recurrence of these traditional algorithms to a significant number of approaches. It makes these strategies such as layer customization, transfer learning, fine-tune, and hyperparameters groups with emphasis on the predominance of the investigated methods.

Still, regarding this relationship, we identified that the approaches that propose new architectures with significant advances regarding the implementation of CNNs sign a trend of investigation present in the studies.

\subsection{Types Crops and Disease Causing Pathogens} \label{S:56}

The results showed that in addition to the approaches that investigate a diverse set of crops, tomatoes together with apples, corn, rice, cucumbers, wheat, grapes, potatoes, and bananas are the ones that concentrate the most significant number of investigations using CNNs. Taking as a concept that grains and cereals are the primary sources of energy ingested by humans, it was evident the lack of studies with approaches that primarily contemplated these crops.

In the case of soybeans, one of the main legumes of industrial interest, it was possible to identify only studies S7, S81, and S84. Highlighted is the approach proposed by \cite{Nagasubramanian:2019} in study S7 that creates an architecture called 3D-CNN that can be used to extract features jointly across the spatial and spectral dimension for classification of a 3D hyperspectral data. The authors demonstrated that a 3D CNN model could be used effectively to learn from hyperspectral data to identify charcoal rot disease in soybean stems. They how to show that saliency map visualization can be used to explain the importance of specific hyperspectral wavelengths in the classification of the diseased and healthy soybean stem tissue.

When analyzing the results that deal with the types of plant diseases, we establish a relationship between the diseases investigated in the approaches with their causative agents. It was possible to observe that the vast majority of investigations identify and classify diseases caused by fungi. This fact can be explained because it is estimated that 70\% of the main plant diseases are caused by this pathogen\cite{Agrios:2005,Godfray:2016}.

The diseases caused by viruses and bacteria, although they do not have consistent quantitative approaches, are represented in a diversity of crops found in the data sets. The diseases caused by algae are represented by the crop Cashew, Citrus, Guava, Strawberries, and tea leaves. We also identified the classification of diseases caused by abiotic factors in investigations using CNNs.

The inexpressive number of approaches that classify diseases caused by nematodes caught our attention. These pathogens are considered to be one of the most significant phytosanitary risks in the main grain and cereal crops in the world. Indeed, the absence of investigations using CNNs to identify symptoms caused by nematodes is the symptomatic similarity of the characteristics presented by the plant with other diseases or particularities that are not visibly expressed in the leaves and stems of the plant.

\section{Conclusion}
\label{S:6}

The automatized plant disease detection systems propose the identification and classification of plant diseases, reconciling the know-how of specialists in phytopathology and the ability to extract symptomatic features, through convolutional neural networks.

The diversity of problems and the specificities of real-world scenarios make it difficult to semantically catalog the data in representative sets with a sufficient number of labeled samples. This problem becomes a relevant and challenging bottleneck to make machine learning methods more applicable in practice.

Our results demonstrate the significant advances in the use of CNN in plant disease prediction processes. It was possible to observe that the traditional architectures combined with optimization and customization methods, despite the complexity of the data set composed with images captured in real environments of the crops, present relevant accuracy. The tendency of the may approaches that propose new CNN architectures based on the process of identifying plant diseases is growing, even presenting lower accuracy than the methods that use traditional architectures.

Finally, we also find that crops such as grains and cereals with high food and financial representativeness are often overlooked by the approaches. Also, we highlight the inexpressive number of methods that identify or classify diseases caused by nematodes.

The study of the works presented and the problems they propose to solve, together with the new trends in architecture of convolutional neural networks, aimed at the identification and classification of plant diseases. These techniques are combined with multispectral and hyperspectral images, which can lead to new solutions for agriculture, bringing production gains with the minimization of damages caused by biotic or abiotic agents.

\newpage
\clearpage

\KOMAoptions{paper=landscape,pagesize}
\newgeometry{top=2cm,textwidth=10cm,textheight=15cm}
\afterpage{\clearpage}
\appendix

\section{Summarization}\label{appendiceA}

\begin{center}

    	\vspace{-.25cm}
		\fontencoding{T1}
		\fontseries{m}
		\fontshape{n}
		\fontsize{4}{7}
		\selectfont 
		\setlength{\tabcolsep}{4pt}
	
\begin{longtable}{rlccccccccc}
		\caption{Characteristics of the Primary Studies of the Final Selection} \label{Tab:finalSelect}
	 \\
		
		
		\hline 
		
		\multicolumn{1}{p{0.015cm}}{\textbf{ID}}                       &
		\multicolumn{1}{l}{\textbf{Refs}}                      & 
		\multicolumn{1}{c}{\textbf{Year}}                      &
		\multicolumn{1}{c}{\textbf{IQ}}                        &
		\multicolumn{1}{c}{\textbf{Algorithm}}         &
		\multicolumn{1}{c}{\textbf{\begin{tabular}[c]{@{}c@{}}Best\\ Accuracy\end{tabular}}}                                 &		    
		\multicolumn{1}{c}{\textbf{Dataset Characteristics}} &
		\multicolumn{1}{c}{\textbf{\begin{tabular}[c]{@{}c@{}}Type of\\ Crops\end{tabular}}}           &
		\multicolumn{1}{c}{\textbf{\begin{tabular}[c]{@{}c@{}}Type of\\  Images\end{tabular}}}                                  &
		
		\multicolumn{1}{c}{\textbf{\begin{tabular}[c]{@{}c@{}}Language \\ \& Framework\end{tabular}}}                                & \multicolumn{1}{c}{\textbf{Predominant Approach}}
		
		\\ \hline
		
		\endfirsthead

		\multicolumn{11}{c}%
		{{\bfseries \tablename\ \thetable{} -- continued from previous page}} \\
		
		\hline 
	
		\multicolumn{1}{p{0.015cm}}{\textbf{ID}}                       &
		\multicolumn{1}{l}{\textbf{Refs}}                      & 
		\multicolumn{1}{c}{\textbf{Year}}                      &
		\multicolumn{1}{c}{\textbf{IQ}}                        &
		\multicolumn{1}{c}{\textbf{Algorithm}}         &
		\multicolumn{1}{c}{\textbf{\begin{tabular}[c]{@{}c@{}}Best\\ Accuracy\end{tabular}}}                                 &		    
		\multicolumn{1}{c}{\textbf{Dataset Characteristics}} &
		\multicolumn{1}{c}{\textbf{\begin{tabular}[c]{@{}c@{}}Type da\\  Crops\end{tabular}}}           &
		\multicolumn{1}{c}{\textbf{\begin{tabular}[c]{@{}c@{}}Type das\\  Images\end{tabular}}}                                  &
		
		\multicolumn{1}{c}{\textbf{\begin{tabular}[c]{@{}c@{}}Linguagem \\ \& Framework\end{tabular}}}                                & \multicolumn{1}{c}{\textbf{Predominant Approach}} \\ \hline
		
		\endhead
		
		\hline \multicolumn{11}{|r|}{{continued from previous page}} \\ \hline
		\endfoot
		
		\hline 
		\endlastfoot
 S1 &
  \cite{ShanwenZhang:2019} &
  2019 &
  5 &
  GPDCNN &
  94,65\% &
  \begin{tabular}[c]{@{}c@{}}Customized\\ 700 Images / 7 Class\\ Real Environment\end{tabular} &
  Cucumbers &
  RGB &
  TensorFlow &
  New Architecture \\  \hline
  S2 &
  \cite{Mehedi:2019} &
  2019 &
  4,5 &
  \begin{tabular}[c]{@{}l@{}}R-CNN \& \\ Faster R-CNN\end{tabular} &
  93,4\% &
  \begin{tabular}[c]{@{}c@{}}Customized\\ 335 Images / 3 Class\\ Real Environment\end{tabular} &
  Wheat &
  RGB &
  TensorFlow &
  Transfer Learning \\ \hline
  S3 &
  \cite{Parul:2019} &
  2019 &
  4,5 &
  S-CNN &
  98,6\% &
  \begin{tabular}[c]{@{}c@{}}PlantVillage Subset\\ 15.817 Images / 8 Class\\ Controlled Environment\\ \\ Customized\\ 17.000 Images / 10 Class\\ Real Environment\end{tabular} &
  Tomato &
  RGB &
  TensorFlow &
  Image Segmentation \\ \hline
  S4 &
  \cite{Liang:2019} &
  2019 &
  4,5 &
  ResNet &
  98\% &
  \begin{tabular}[c]{@{}c@{}}PlantVillage\\ 32.933 Images / 45 Class\\ Controlled Environment\end{tabular} &
  Diverse &
  RGB &
  PyTorch &
  \begin{tabular}[c]{@{}c@{}}Fine Tune \&\\ Hyperparameters\end{tabular} \\ \hline
S5 &
  \cite{HuangLiu:2019} &
  2019 &
  4,5 &
  UNet + InceptionV3 &
  87,45\% &
  \begin{tabular}[c]{@{}c@{}}Crop Disease AI Challenger 2018\\ 40.000 Images / 19 Class\\ Controlled Environment\end{tabular} &
  Diverse &
  RGB &
  TensorFlow &
  Image Segmentation \\ \hline
S6 &
  \cite{ArtzaiPicon:2019} &
  2019 &
  4 &
  RESNET-MC1 &
  \begin{tabular}[c]{@{}c@{}}98\%\\ BAC\end{tabular} &
  \begin{tabular}[c]{@{}c@{}}Customized\\ 121.955 Images / 21 Class\\ Controlled Environment e Real\end{tabular} &
  Diverse &
  RGB &
  TensorFlow/Keras &
  New Architecture \\ \hline
S7 &
  \cite{Nagasubramanian:2019} &
  2019 &
  4 &
  3D-CNN &
  95,73\% &
  \begin{tabular}[c]{@{}c@{}}Customized\\ 1.090 Images / 2 Class\\ Controlled Environment\end{tabular} &
  Soybean &
  Hyperspectral &
  TensorFlow/Keras &
  New Architecture \\
\end{longtable}
	\vspace{-0.2cm}

\end{center}

\newpage
\KOMAoptions{paper=portrait,pagesize}
\recalctypearea
\section{Summarization}\label{appendiceB}


\begin{center}
    
		\vspace{-.25cm}
		\fontencoding{T1}
		\fontsize{4}{6}

		\selectfont 
\begin{longtable}[c]{cccll}
 \caption{Diseases and their Pathogens grouped by Dataset and Studies}
 \label{Tab:PathogenStudies}\\
\hline
\textbf{ID Study} &
  \textbf{DataSet Name} &
  \textbf{Type Crop} &
  \textbf{Disease Name} &
  \textbf{Pathogen} \\ \hline
\endfirsthead
\multicolumn{5}{c}%
{{\bfseries Table \thetable\ continued from previous page}} \\
\hline
\textbf{ID Study} &
  \textbf{DataSet Name} &
  \textbf{Type Crop} &
  \textbf{Disease Name} &
  \textbf{Pathogen} \\ \hline
\endhead
\multirow{34}{*}{\begin{tabular}[c]{@{}c@{}}S3, S4, S9, S11,\\  S13, S14, S18,\\  S21, S23, S26,\\  S27, S28, S29, \\ S35, S41, S48,\\  S51, S53, S55,\\  S56, S57, S60, \\ S61, S62, S63,\\  S65, S66, S67,\\  S68, S69, S72,\\  S73, S76, S78,\\  S81, S85, S86,\\  S93, S99, S101,\\  S105, S108, S110,\\  S114, S117\end{tabular}} &
  \multirow{34}{*}{PlantVillage} &
  \multirow{4}{*}{Apple} &
  Apple scab &
  Fungi \\ \cline{4-5} 
 &
   &
   &
  Black rot &
  Fungi \\ \cline{4-5} 
 &
   &
   &
  Cedar apple rust &
  Fungi \\ \cline{4-5} 
 &
   &
   &
  Leaf Black Rot &
  Fungi \\ \cline{3-5} 
 &
   &
  Cherry &
  Powdery mildew &
  Fungi \\ \cline{3-5} 
 &
   &
  \multirow{4}{*}{Corn} &
  Cercospora leaf spot &
  Fungi \\ \cline{4-5} 
 &
   &
   &
  Gray leaf spot &
  Fungi \\ \cline{4-5} 
 &
   &
   &
  Common rust &
  Fungi \\
\hline
\end{longtable}
\end{center}

\vskip3pt

\vskip5pt

\noindent Andre S. Abade and Fl{\'a}vio de Barros Vidal (\textit{University of Bras{\'i}lia, Bras{\'i}lia, Brazil} and Paulo Afonso Ferreira (Department of Agronomy, Federal University of Mato Grosso, Brazil)
\vskip3pt

\noindent E-mail: fbvidal@unb.br

\bibliographystyle{unsrt}  
\bibliography{referencias}

\end{document}